\documentclass{article}
\usepackage{spconf,amsmath,graphicx}
\usepackage{booktabs} 
\usepackage{amsfonts}
\usepackage{multirow}
\usepackage{multicol}
\usepackage{makecell}

\title{A parameterized generative adversarial network using cyclic projection for explainable medical image classifications}

\name{%
\begin{tabular}{@{}c@{}}
Xiangyu Xiong$^{1}$, Yue Sun$^{1}$, Xiaohong Liu$^{2}$, Chan-Tong Lam$^{1}$, Tong Tong$^{3}$, Hao Chen$^{4}$ \\
Qinquan Gao$^{3}$, Wei Ke$^{1}$, Tao Tan$^{1, \star}$\thanks{$^{\star}$Corresponding author}
\end{tabular}}

\address{$^{1}$ Faculty of Applied Sciences, Macao Polytechnic University \\
         $^{2}$ John Hopcroft Center (JHC) for Computer Science, Shanghai Jiao Tong University \\
         $^{3}$ College of Physics and Information Engineering, Fuzhou University\\
         $^{4}$ Department of Mathware, Jiangsu JITRI Sioux Technologies Co., Ltd}

\begin{document}
%\ninept
%
\maketitle
\begin{abstract}
Although current data augmentation methods are successful to alleviate the data insufficiency, conventional augmentation are primarily intra-domain while advanced generative adversarial networks (GANs) generate images remaining uncertain, particularly in small-scale datasets. In this paper, we propose a parameterized GAN (ParaGAN) that effectively controls the changes of synthetic samples among domains and highlights the attention regions for downstream classification. Specifically, ParaGAN incorporates projection distance parameters in cyclic projection and projects the source images to the decision boundary to obtain the class-difference maps. Our experiments show that ParaGAN can consistently outperform the existing augmentation methods with explainable classification on two small-scale medical datasets.
\end{abstract}
\section{Introduction}
Deep neural networks have achieved success in computer vision fields \cite{DLreview}, where a large-scale dataset is crucial for effectively training. However, training neural networks on the small-scale datasets leads to the overfitting and poor generalization. Although regularization techniques \cite{dropout, batchnormlization, layernormalization, groupnormalization} have been developed to prevent overfitting, augmenting training data is an effective option to address the data insufficiency.

Researchers have developed two main data augmentation methods in the past decade. Conventional augmentation method can explore prior knowledge \cite{KrizhevskySH12, High-Performance}, including random cropping, flipping, etc. However, its intra-domain augmented data contributes little to the description of decision boundaries for downstream classifications. Generative adversarial network (GAN) \cite{GAN} and its variations have also emerged as augmentation means \cite{DCGAN, liver, matrix}. However, the synthetic images are quality uncertain and variety insufficiency because of the mode collapsing \cite{ModeCollapse}. Moreover, there is still uncertainty regarding the domain labels of the synthetic images even if soft labels are used~\cite{HongjiangShi, ECGAN}.

Auxiliary information have been integrated with GANs to synthesize images of a given specific type. cGAN~\cite{cGAN} generates images with a specific condition of class labels. ACGAN~\cite{ACGAN} and VACGAN~\cite{VACGAN} introduce an auxiliary classifier to reconstruct the class labels. Later, this idea is expanded to cross-domain translation by reconstructing target domain labels, such as cCycleGAN~\cite{AGcCyeleGAN, cCycleGAN} and StarGAN~\cite{stargan}. However, the class label and domain label cannot make synthetic samples adjust the decision boundary for classifications.

To overcome these issues, we propose a parameterized GAN (ParaGAN) enabling controlling the degree variation of synthetic images. This is achieved by leveraging target-domain samples' distances to the optimal hyperplane as controllable parameters in forward path and considering source images' distances to the hyperplane in the backward reconstruction path. Experiments show that ParaGAN consistently outperforms the state-of-the-arts and provides a more transparent explanation than Grad-CAM \cite{gradcam}.

Our main contributions are as follows: 1. A novel cyclic parameterized projection perpendicular to hyperplane controlling the variety of synthetic images. 2. An online augmentation manner with weighted synthetic loss enhancing binary classification. 3. A novel class-difference map enabling explaining the downstream classification.

\begin{figure*}[tbp]
	\centering
	\includegraphics[width=0.90\textwidth]{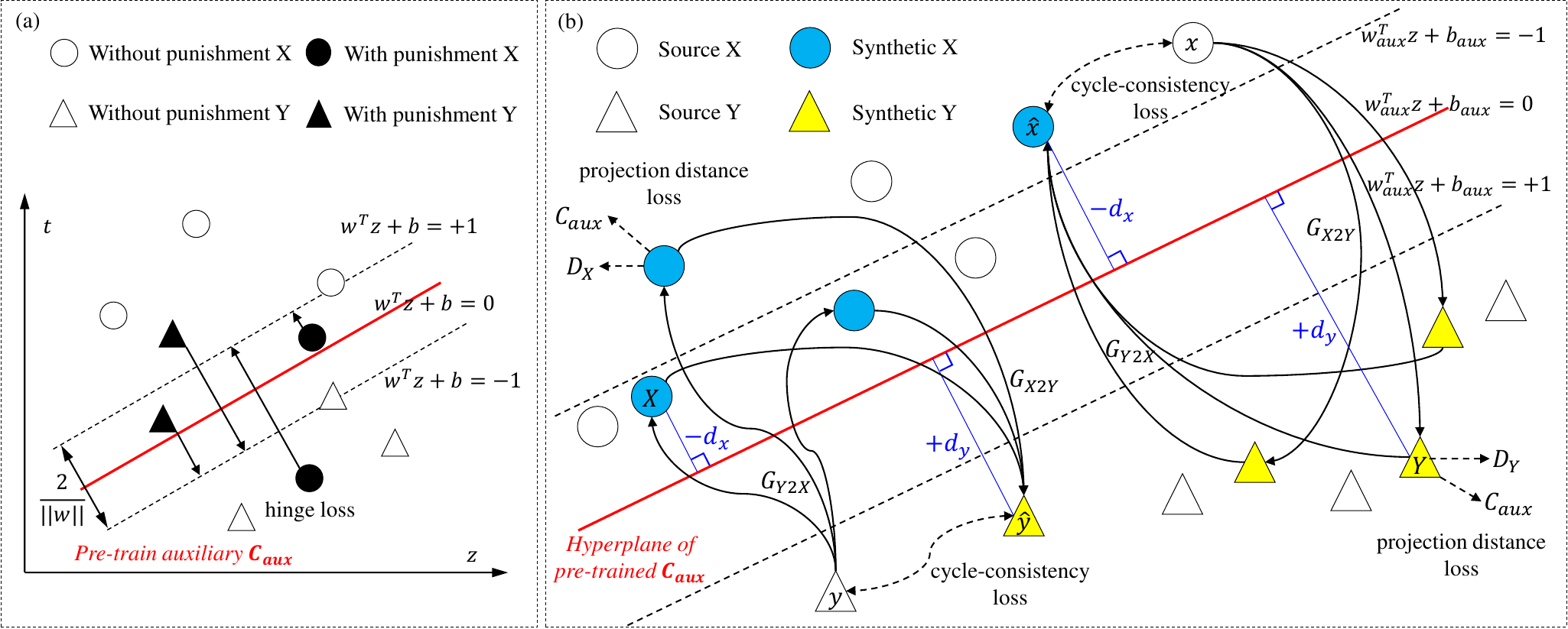}
	\caption{Overview of ParaGAN enabling cyclic parameterized projection. (a) Pre-train an auxiliary classifier $C_{aux}$ by hinge loss to provide a hyperplane. (b)  The generators translate source images conditioned on the target images' projection distances in forward path, and vice versa for reconstructing sources. $C_{aux}$ reconstructs the projection distances from synthetic images.}
	\label{Architecture}
\end{figure*}

\section{Materials and Methods}

\subsection{Dataset Acquisition and Evaluation Metrics}
We collect a mixed breast ultrasound datasets which contains BUSI \cite{BUSI} and UDIAT \cite{UDIAT} and a COVID-19 Dataset (COVID-CT) \cite{zhao2020COVID-CT-Dataset}. Accuracy (ACC) and the area under the receiver operating characteristic curve are used for evaluation.

\subsection{Parameterized Generative Adversarial Network}
We pre-train a binary classifier by hinge loss to obtain an optimal hyperplane dividing the two domain samples, as shown in Figure \ref{Architecture}(a). For a linear binary classifier $ t = w^Tz+b $, given a training set $\left\{ z_i \right\}_{i=1}^N$, $z_i \in R^D$, $t_i \in \left\{ -1, +1 \right\}$, the hinge loss is defined as follows:
\begin{equation}
\small
 \mathcal{L}_{\rm hinge}(t, \hat{t}) = \frac{1}{N}\sum_{i=1}^N max[0, 1-t_i(w^Tz_i + b))],
\end{equation}
We develop parameterized generators $G_{X2Y}$ and $G_{Y2X}$ taking the projection distances to hyperplane as input parameters. To achieve this, we use pre-trained $C_{aux}$ to measure projection distances from the real images to the optimal hyperplane ($w_{aux}^Tz+b_{aux}=0$). Given two random real images $x$ and $y$, their vertical projection distances $d_x$ and $d_y$ are as:
\begin{equation}
\small
d_x = \lvert C_{aux}(x) \rvert, x\in X; d_y = \lvert C_{aux}(y) \rvert, y\in Y.
\end{equation}

Figure \ref{Architecture}(b) illustrates the synthesis phase. Firstly, $G_{X2Y}$ translates a source image $x$ into $G_{X2Y}(x, +d_y)$ and $D_Y$ distinguishes between this translated image and the real image $y$, and vice versa for $G_{Y2X}$ and $D_X$. Secondly, an auxiliary classifier $C_{aux}$ reconstructs the vertical projection distances $d_y$ and $d_x$ from the translated images $G_{X2Y}(x, +d_y)$ and $G_{Y2X}(y, -d_x)$, respectively. Thirdly, the source image $x$ is reconstructed from the translated image $G_{X2Y}(x, +d_y)$ using its projection distance $d_x$, and vice versa for source $y$.
\begin{figure*}[tbp]
	\centering
	\includegraphics[width=0.95\textwidth]{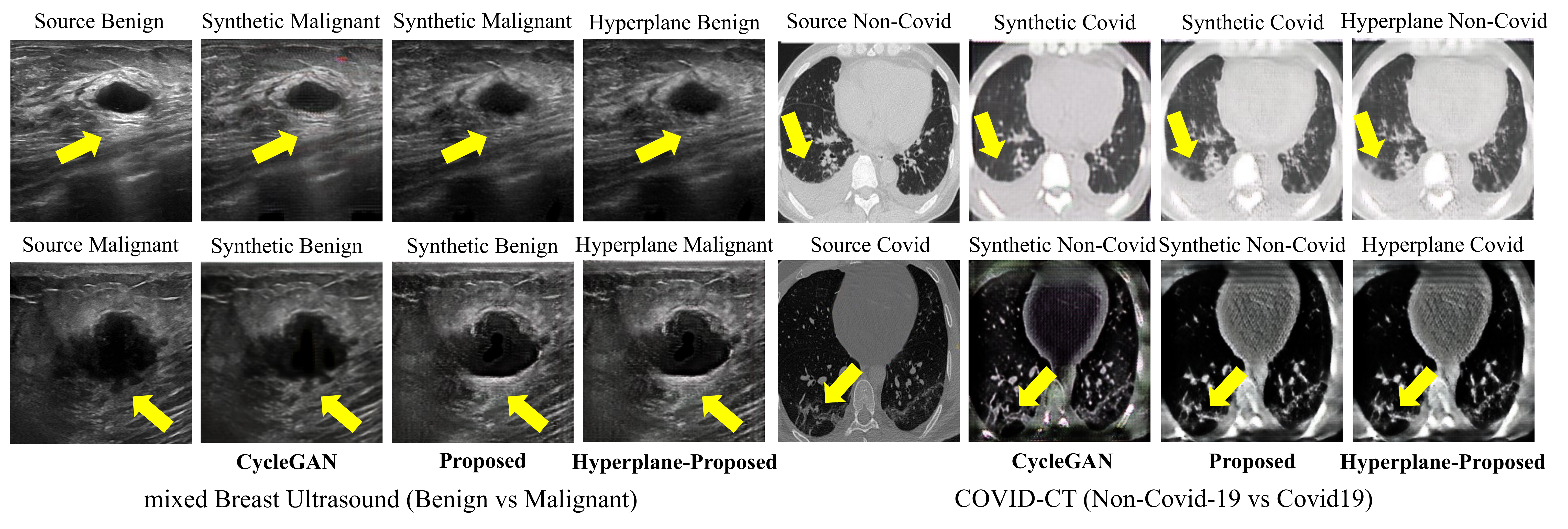}
	\caption{Qualitative results over mixed Breast Ultrasound and COVID-CT. The synthetic images by the ParaGAN are clearly closer to target domain than that by the CycleGAN \cite{cyclegan}. The projected results on the hyperplane for source images are displayed.}
	\label{Qualitative}
\end{figure*}

\subsection{Objective Function}
In the proposed ParaGAN, the adversarial loss for $G_{X2Y}$ and $D_Y$ is expressed as:
\begin{equation}
\small
\begin{split}
\mathcal{L}_{\rm GAN}(G_{X2Y}, &D_Y, X, Y) = \mathbb{E}_{y\in Y}[logD_Y(y)]\\& + \mathbb{E}_{x\in X}[log(1 - D_Y(G_{X2Y}(x, +d_y))],
\end{split}
\end{equation}
and vice versa for $G_{Y2X}$ and $D_X$.

We propose a projection distance loss to force the synthetic images having same vertical distances as the target images. The projection distance loss can be formulated as:
\begin{equation}
\small
\begin{split}
\mathcal{L}_{\rm proj}(&G_{X2Y}, G_{Y2X}, C_{aux}) \\& = \mathbb{E}_{x\in X}[||(+d_y) - C_{aux}(G_{X2Y}(x, +d_y))||_2^2] \\& + \mathbb{E}_{y\in Y}[||(-d_x) - C_{aux}(G_{Y2X}(y, -d_x))||_2^2].
\end{split}
\end{equation}

Cycle consistency loss is introduced to establish relationships between individual input $x_i$ and a desired output $y_i$:
\begin{equation}
\small
\begin{split}
\mathcal{L}_{\rm cyc}(& G_{X2Y}, G_{Y2X}) \\& = \mathbb{E}_{x\in X}[||x - G_{Y2X}(G_{X2Y}(x, +d_y), -d_x)||_1] \\& + \mathbb{E}_{y\in Y}[||y -G_{X2Y}(G_{Y2X}(y, -d_x), +d_y)||_1].
\end{split}
\end{equation}

Finally, the objective function for cross-domain synthesis represented by the following equation:
\begin{equation}
\small
\begin{split}
&\mathcal{L}_{\rm ParaGAN}(G_{X2Y}, G_{Y2X}, C_{aux}, D_X, D_Y) \\& = \mathcal{L}_{\rm GAN}(G_{X2Y}, D_Y, X, Y) + \mathcal{L}_{\rm GAN}(G_{Y2X}, D_X, Y, X) \\& + \lambda_{\rm proj}\mathcal{L}_{\rm proj}(G_{X2Y}, G_{Y2X}, C_{aux}) \\& + \lambda_{\rm cyc}\mathcal{L}_{\rm cyc}(G_{X2Y}, G_{Y2X}).
\end{split}
\end{equation}
where $\lambda_{\rm proj}$ and $\lambda_{\rm cyc}$ are weights that control the relative importance of projection distance loss and cycle consistency loss, respectively, compared to the adversarial loss.

\subsection{Downstream Classifier Optimization}
The loss of synthetic images in downstream classification is multiplied by a hyperparameter $\alpha$, which controls the relative importance of synthetic data compared to true images. The combined loss of the downstream classifier can be defined as:
\begin{equation}
\small
\mathcal{L}_{\rm C}(c, \hat{c}) =  \mathcal{L}_{\rm hinge}(c, \hat{c}_{real}) + \alpha\mathcal{L}_{\rm hinge}(c, \hat{c}_{syn}).
\end{equation}
Where $c$ is the respective labels, $\hat{c}_{real}$ and $\hat{c}_{syn}$ are the downstream classifier's outputs for real data and synthetic data respectively, $\alpha$ is the loss weight of the synthetic data.
\begin{table}[tbp]
  \centering
  \caption{Comparison with the state of the arts over mixed Breast Ultrasound and COVID-CT. The proposed ParaGAN outperforms the state-of-the-arts using limited clinical data.}
  \resizebox{\linewidth}{!}{
    \begin{tabular}{cccccc}
    \toprule
    \multirow{2}[4]{*}{Methods} & \multirow{2}[4]{*}{Loss} & \multicolumn{2}{c}{mixed Breast Ultrasound} & \multicolumn{2}{c}{COVID-CT} \\
\cmidrule{3-6}          &       & ACC   & AUC   & ACC   & AUC \\
    \midrule
    Original & CrossEntropy & 0.849$\pm$0.000     & 0.928$\pm$0.013     & 0.767$\pm$0.049     & 0.836$\pm$0.043 \\
    \midrule
    Original & HingeLoss & 0.887$\pm$0.025     & 0.933$\pm$0.019     & 0.757$\pm$0.079     & 0.840$\pm$0.071 \\
    \midrule
    \makecell{Conventional \\ Augmentation (CA)}    & CrossEntropy & 0.877$\pm$0.016     & 0.933$\pm$0.003     & 0.765$\pm$0.015     & 0.845$\pm$0.027  \\
    \midrule
    \makecell{Conventional \\ Augmentation (CA)}    & HingeLoss & 0.862$\pm$0.030     & 0.934$\pm$0.013     & \textbf{0.796$\pm$0.010}     & 0.874$\pm$0.013 \\
    \midrule
    CA + ACGAN & HingeLoss & 0.824$\pm$0.005     & 0.916$\pm$0.010     & 0.777$\pm$0.031     & 0.863$\pm$0.012 \\
    \midrule
    CA + VACGAN & HingeLoss & 0.843$\pm$0.005     & 0.929$\pm$0.010     & 0.772$\pm$0.032     & 0.829$\pm$0.039 \\
    \midrule
    CA + CycleGAN & HingeLoss & 0.849$\pm$0.028     & 0.915$\pm$0.003     & 0.785$\pm$0.017     & 0.866$\pm$0.017 \\
    \midrule
    CA + Proposed & HingeLoss & \textbf{0.895$\pm$0.048}     & \textbf{0.947$\pm$0.017}     & \textbf{0.796$\pm$0.003}     & \textbf{0.883$\pm$0.027} \\
    \bottomrule
    \end{tabular}%
    }
  \label{tab1}%
\end{table}%

\begin{table}[tbp]
  \centering
  \caption{Comparison with weighted synthetic losses ($\alpha$) over mixed Breast Ultrasound and COVID-CT. Table shows optimal values of $\alpha$ are 0.2 and 1.0 for two datasets.}
  \resizebox{\linewidth}{!}{
    \begin{tabular}{ccccc}
    \toprule
    \multirow{2}[4]{*}{ \makecell{weighted synthetic loss $\alpha$}} & \multicolumn{2}{c}{mixed Breast Ultrasound} & \multicolumn{2}{c}{COVID-CT} \\
\cmidrule{2-5}          & ACC   & AUC   & ACC   & AUC \\
    \midrule
    0.2 & \textbf{0.895$\pm$0.048}     & \textbf{0.947$\pm$0.017}     & 0.793$\pm$0.021     & 0.874$\pm$0.021 \\
    \midrule
    0.4 & 0.871$\pm$0.011     & 0.938$\pm$0.020     & 0.777$\pm$0.017     & 0.880$\pm$0.005 \\
    \midrule
    0.6 & 0.855$\pm$0.022     & 0.936$\pm$0.014     & 0.782$\pm$0.012     & 0.861$\pm$0.002 \\
    \midrule
    0.8 & 0.862$\pm$0.038     & 0.938$\pm$0.008     & \textbf{0.806$\pm$0.020}     & 0.858$\pm$0.021 \\
    \midrule
    1.0 & 0.846$\pm$0.005     & 0.940$\pm$0.016     & 0.796$\pm$0.003     & \textbf{0.883$\pm$0.027} \\
    \bottomrule
    \end{tabular}%
     }
  \label{tab2}%
\end{table}%

\section{Implementation}
\subsection{Network Architecture}
The generator's architecture is adopted from CycleGAN \cite{cyclegan}. We add one channel for the first convolutional layer, because the projection distance is required to spatially replicated match the size of the input image and concatenated with the input image. We use PatchGANs \cite{pix2pix} as discriminator network to detect whether $70\times 70$ overlapping image patches are real or synthetic. We adopt ConvNeXt \cite{convnext} for the auxiliary classifier and the downstream classifier.

\subsection{Training Settings}
We train all networks with a learning rate of 0.0002 for the first 25 epochs and linearly decay the learning rate to 0 over the next 25 epochs. The hyperparameter $\lambda_{\rm cyc}$ is set to 10, and $\lambda_{\rm proj}$ is set to 0.1 considering the large scale of the distances. We adopt transfer learning for training all classifiers using ImageNet Dataset. The auxiliary classifier $C_{aux}$ is not updated during the ParaGAN training procedure. We report ACC and AUC averaged over three runs.
\begin{figure*}[htbp]
	\centering
	\includegraphics[width=0.90\textwidth]{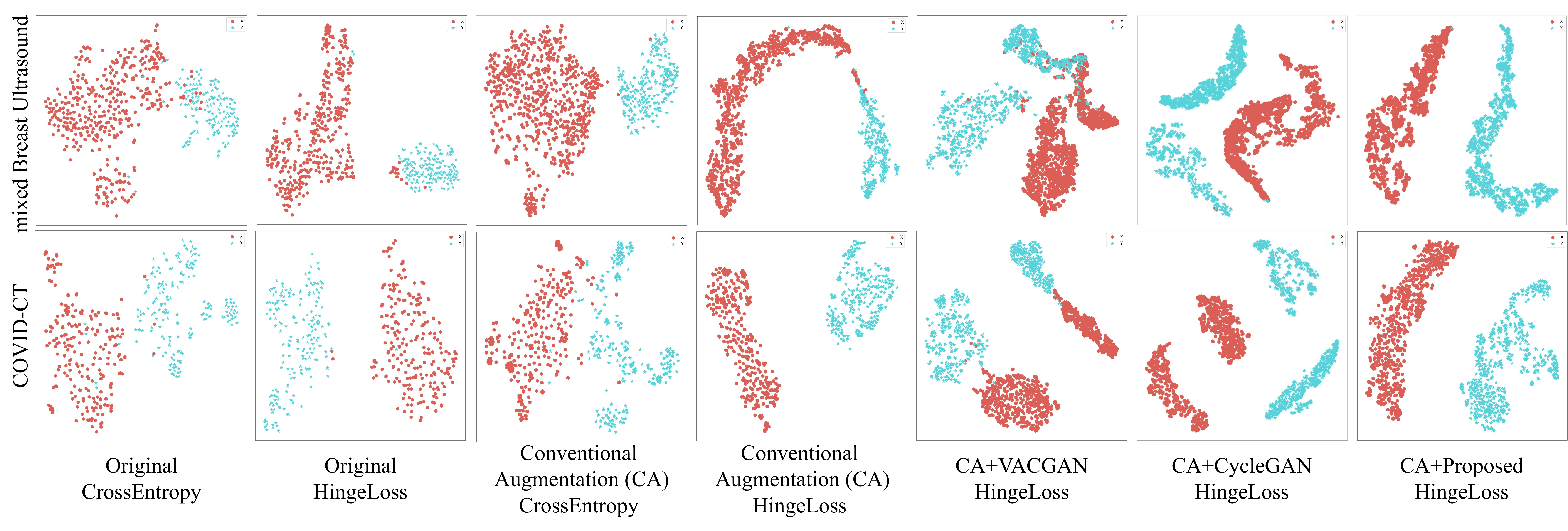}
	\caption{Training samples distributions of various methods for downstream ConvNext visualized by t-SNE \cite{van2008visualizing}. The proposed ParaGAN tends to fill the latent space and to be along the margins of the hyperplane.}
	\label{Distribution}
\end{figure*}
\begin{figure}[htbp]
	\centering
	\includegraphics[width=0.92\columnwidth]{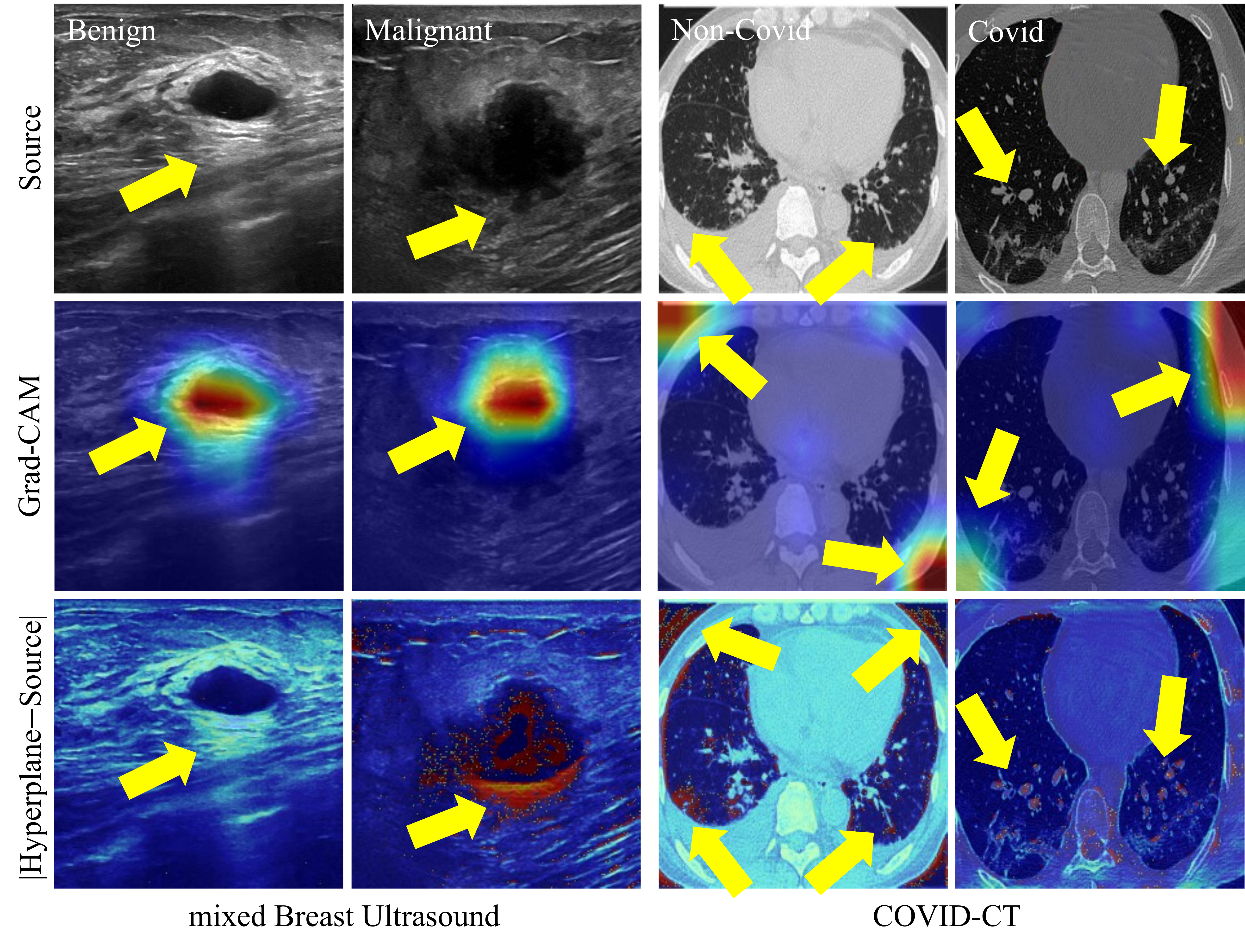}
	\caption{The difference between the source images and their projections on a hyperplane can highlight class-specific region, which cannot be deduced from the Grad-CAM \cite{gradcam}.}
	\label{ClassDifferenceMaps}
\end{figure}

\section{Experiments and Results}
\subsection{Experiments on Breast Ultrasound and COVID-CT}
Table \ref{tab1} quantitatively compares the proposed ParaGAN with several state-of-the-art GANs over datasets mixed Breast Ultrasound and COVID-CT. We can see that the proposed ParaGAN performs the best consistently especially when training samples are limited.

Figure \ref{Qualitative} demonstrates that the proposed ParaGAN can generate more realistic and diverse images compared with the CycleGAN. We can see that the proposed ParaGAN makes the cross-domain changes including posterior regions of lesions in breast ultrasound image and the ground glass in the lungs CT image. Moreover, the source images and their projected images on hyperplane (Hyperplane-Proposed) are used to interpret the downstream classification in Section \ref{Interpretability}.

\subsection{Ablation Study}
The accuracy and auc can increase considerably by adjusting the synthetic weight (Table \ref{tab2}). With synthetic weights of 0.2 and 1.0, the auc of downstream classifier achieve the best performance. Using a synthetic weight allows for effective usage of GAN synthesized images, as synthetic images generally will not be as beneficial for a classifier as real images because of their faulty and somewhat unreliable nature.

\subsection{Distribution of Training Samples}
Figure \ref{Distribution} illustrates that how the cross-domain samples could help shape the decision boundary in the latent space from a downstream classifier. VACGAN converts noise-vector to images with condition of binary class labels, and CycleGAN translates source image to target image with only image space constraint. These two methods lead to the label uncertainty of synthetic samples. The proposed ParaGAN translates source images to target images with both the image space constraint and the hyperplane distance constraint.

\subsection{Interpretability of Classification}
\label{Interpretability}
Figure \ref{ClassDifferenceMaps} shows that ParaGAN provides the class-difference maps (CDMs) for explaining the binary classifier. The Grad-CAM focuses on the regions with large gradient changes in each class image, whereas the proposed class-difference maps (CDMs) focuses regions with large changes between each image and its projection on hyperplane. Specifically, we define the CDMs for images in domain $X$ and domain $Y$ using $|x - G_{X2Y}(x, 0))|$ and $|y - G_{Y2X}(y, 0))|$ respectively.

\section{Conclusion and Future Works}
We show that ParaGAN can generate samples to benefit and explain the downstream classification tasks, especially in small-scale datasets. We observe that controlling synthetic samples' variety have significantly more impact than blindly augmentation, and difference between the images and their projections on decision boundary can contribute explanation.

The current work has limitations that need to be studied in future. 1. Our work primarily focuses on the binary classification, and thus we will investigate levering hyperplane among multi-classes. 2. We will conduct a full-spectrum evaluation of the synthetic images in terms of clinical usefulness.

\section*{Acknowledgments}
This work is supported in part by the Macao Polytechnic University Grant (RP/FCA-05/2022) and in part by the Science and Technology Development Fund of Macao (0041/2023/RIB2).

\bibliographystyle{IEEEbib}
\bibliography{ParaGAN}

\end{document}